\documentclass[journal]{IEEEtran}
\usepackage{generic}
\usepackage{cite}
\usepackage{amsmath,amssymb,amsfonts}
\usepackage{algorithmic}
\usepackage{graphicx}
\usepackage{textcomp}

\usepackage{hyperref}
\usepackage{subcaption}
\usepackage{array, booktabs}
\usepackage{balance}
\newcolumntype{P}[1]{>{\centering\arraybackslash}p{#1}}
\newcolumntype{M}[1]{>{\centering\arraybackslash}m{#1}}
\usepackage{adjustbox}
\usepackage{multirow}
\usepackage{caption}
\usepackage{float}
\usepackage{threeparttable}

\def\BibTeX{{\rm B\kern-.05em{\sc i\kern-.025em b}\kern-.08em
    T\kern-.1667em\lower.7ex\hbox{E}\kern-.125emX}}
\newcommand{\etal}{\emph{et~al.}}

\begin{document}

\title{A Novel Hierarchical-Classification-Block Based Convolutional Neural Network for Source Camera Model Identification}

\author{Mohammad Zunaed, Shaikh Anowarul Fattah, \IEEEmembership{Senior Member, IEEE}
\thanks{Mohammad Zunaed and Shaikh Anowarul Fattah are with the Department of Electrical and Electronic Engineering, Bangladesh University of Engineering and Technology, Dhaka-1205, Bangladesh (e-mail: rafizunaed@gmail.com, fattah@eee.buet.ac.bd).}
}

\maketitle

\begin{abstract}
Digital security has been an active area of research interest due to the rapid adaptation of internet infrastructure, the increasing popularity of social media, and digital cameras. Source camera model identification is of prime importance for digital image forensics, especially for law enforcement agencies investigating content manipulation, identity forgery, child sexual abuse material, etc. Due to inherent differences in working principles to generate an image, different camera brands left behind different intrinsic processing noises which can be used to identify the camera brand. In the last decade, many signal processing and deep learning-based methods have been proposed to identify and isolate this noise from the scene details in an image to detect the source camera brand. One prominent solution is to utilize a hierarchical classification system rather than the traditional single-classifier approach. Different individual networks are used for brand-level and model-level source camera identification. This approach allows for better scaling and requires minimal modifications for adding a new camera brand/model to the solution. However, using different full-fledged networks for both brand and model-level classification substantially increases memory consumption and training complexity. Moreover, extracted low-level features from the different network's initial layers often coincide, resulting in redundant weights. To mitigate the training and memory complexity, we propose a classifier-block-level hierarchical system instead of a network-level one for source camera model classification. Our proposed approach not only results in significantly fewer parameters but also retains the capability to add a new camera model with minimal modification. 
Thorough experimentation on the publicly available Dresden dataset shows that our proposed approach can achieve the same level of state-of-the-art performance but requires fewer parameters compared to a state-of-the-art network-level hierarchical-based system. 
\end{abstract}

\begin{IEEEkeywords}
Source camera model identification, digital image forensics, sensor pattern noise.
\end{IEEEkeywords}

\section{Introduction}
\IEEEPARstart{S}{ource} camera model identification (SCMI) has been an active area of research interest for forensic analysis to investigate digital image and video image information forgery \cite{6515027}. Due to the rapid growth of internet infrastructures and the increasing availability of media consumption devices, digital images are used more frequently in everyday life. Unfortunately, this has also led to an increase in the circulation of illicit content and altered images, raising significant security issues. The SCMI can play a vital role in the forensic investigations of these digital images. We can employ SCMI detection tools in various forensic applications, such as image forgery detection, fighting against fake news, and image integrity verification \cite{8014966, 8713484, Li_1, 6987281}. \par
Several methods have been proposed over the last decade for identification of a camera model. An extensive review of these methods can be found in \cite{6515027, nikon_merge}. Metadata from the Exchangeable Image File (EXIF) header can be used for identifying clues regarding the source camera. However, they can easily be altered or removed, making them unreliable for the task of SCMI. In contrast, it has been shown in the research that image tampering can be detected from the pixel values even if the image is altered. For example, it is possible to identify some common image processing operations such as median filtering \cite{6558797}, Gaussian filtering \cite{7368606, 4722494}, and JPEG image compression \cite{4722494}, among others. Therefore, SCMI algorithms that analyze pixel values are more robust compared to Exchangeable Image File (EXIF) header. \par
As smartphones increasingly become a daily commodity, image editing applications are gaining more popularity. Image manipulation operations such as image resizing, filtering, and compression, among others, not only distort the sensor pattern noise but also leave behind operation-specific traces in the images. It makes the task of camera model identification even more challenging. While the presence of these additional traces makes the task of SCMI more difficult, it is still possible to detect their presence \cite{Wang2016, 7368606, 4722494}. Image forgery detection is a crucial forensic task but falls beyond the scope of this work, as we consider only unedited images for this study.\par
Due to recent advances in deep learning, many data-driven deep-learning approaches have been adopted over time for source camera model identification. Bayer and Stamm \etal \cite{bayer_stamm} proposed a convolutional neural network with a constrained high-pass filter-based preprocessor to suppress the effect of image edges and textures on classification performance. Rafi \etal \cite{Rafi2021} proposed a data-driven preprocessing block named remnant block that can learn to filter out scene details for robust classification. However, these models are not flexible for adding a new camera model for a particular brand. To address this, Bennabhaktula \etal \cite{BENNABHAKTULA2022117769} adopted a hierarchical classification approach, where they performed brand classification in the first stage, followed by the classification of camera models in the second stage. In addition, they extracted homogeneous patches from an image and fed them to the neural network for feature extraction. The homogeneous patches are rich in-camera processing noise, which prevents the model from learning scene details. However, individual models for both brand-level and model-level increase training cost and complexity greatly. These issues call for a framework that can utilize the hierarchical scheme in a way that does not substantially increase training cost and the number of network parameters but retains the same level of classification performance as before. \par
It is necessary to clarify what is meant by camera-brand and camera-model identification. In brand identification, the task is to identify the specific camera brand used to capture an image (Sony, Canon, FujiFilm, etc). The model identification task refers to the task that identifies which model of that particular brand is used to capture that image (e.g., Sony\_DSC-H50, Sony\_DSC-T77, Sony\_DSC-W170 models for the Sony brand).\par
In this work, we propose a classifier-block-level hierarchical convolutional neural network system instead of a network-level one for both source camera brand and model classification. Our approach utilizes a single network instead of a multitude of networks, which results in significantly fewer parameters and less training complexity. Our proposed framework is flexible for adding a new camera brand/model with minimal modification. 
The contributions of this paper are summarized as follows:
\begin{itemize}
    \item We propose a novel classifier-block-level hierarchical (HCB) module that facilitates brand and model-level classifications in a united framework in a hierarchical manner. This module significantly reduces the number of network parameters.
    \item Our proposed HCB module allows for easier integration of new camera brands and models into the already trained network.
    \item Our extensive experiments on the publicly available Dresden dataset show that our proposed approach can maintain state-of-the-art performance compared to a network-level hierarchical-based system while requiring fewer parameters and reducing training complexity.
\end{itemize}
The rest of the paper is organized as follows. Section II reviews the related works on source camera model identification. Next, the proposed hierarchical network, along with patch selection and data balancing scheme, are described in Section III. The dataset description and training schemes are presented in Section IV. The results and analysis of comprehensive experiments and proposed hierarchically structured modules are given in Section V. Section V also highlights the possible future research direction. Finally, Section VI concludes the whole work.

\section{Related Works}
\begin{figure}[!t]
	\centering
	\includegraphics[width=\linewidth]{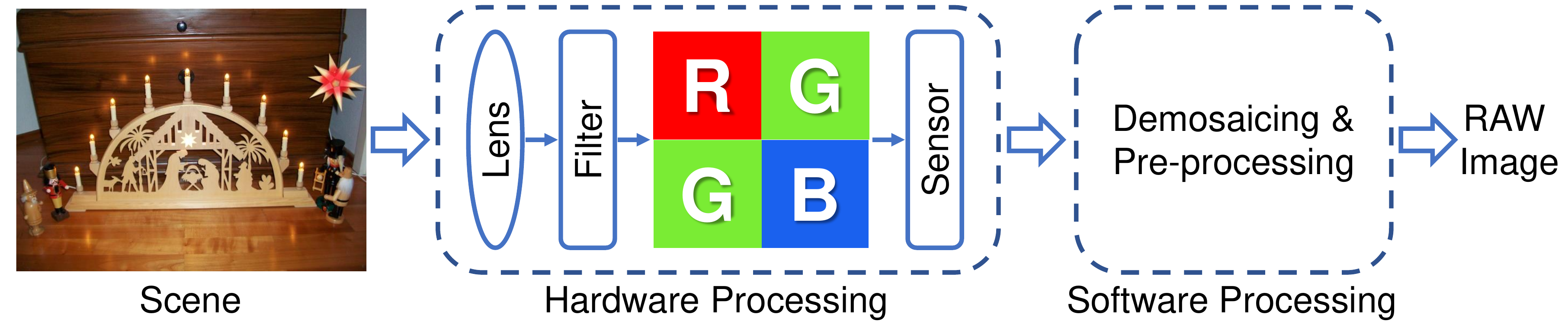}
 	\caption{A high-level image generation pipeline inside a digital camera.}
 	\label{generalized_pipeline_for_image_capture}
\end{figure}

Since digital cameras have become more mainstream, source camera identification has been an active area of research for forensic analysis of digital media. Camera model identification from images is possible due to different types of artifact traits left by different camera brands due to their inherent difference in the image generation pipeline. An illustration of a high-level image generation pipeline inside a digital camera is shown in Fig. \ref{generalized_pipeline_for_image_capture}. The process can be divided into a sequence of hardware and software pipelines. In the hardware pipeline, the light rays from a scene are collected by a set of lenses and passed through anti-aliasing filters, color filter arrays (CFA), and finally, the imaging sensor. The sensor produces the RAW image by converting the analog signal to digital. The software pipeline enhances and generates the final image by applying several preprocessing algorithms such as demosaicing, gamma correction, compression, and other operations. The exact implementation of hardware and software pipelines varies from brand to brand. These operational variations left unique camera traces in the final image, which can be used to identify the camera brand. Even two camera devices of the same brand can leave behind different processing noises due to the unique random noise generated by each imaging sensor in every device. In the last decade, many signal processing and deep-learning based algorithms have been proposed to isolate these in-camera processing noises from the scene details to identify the camera model's brand. 

\begin{figure*}[!t]
	\centering
	\includegraphics[width=\linewidth]{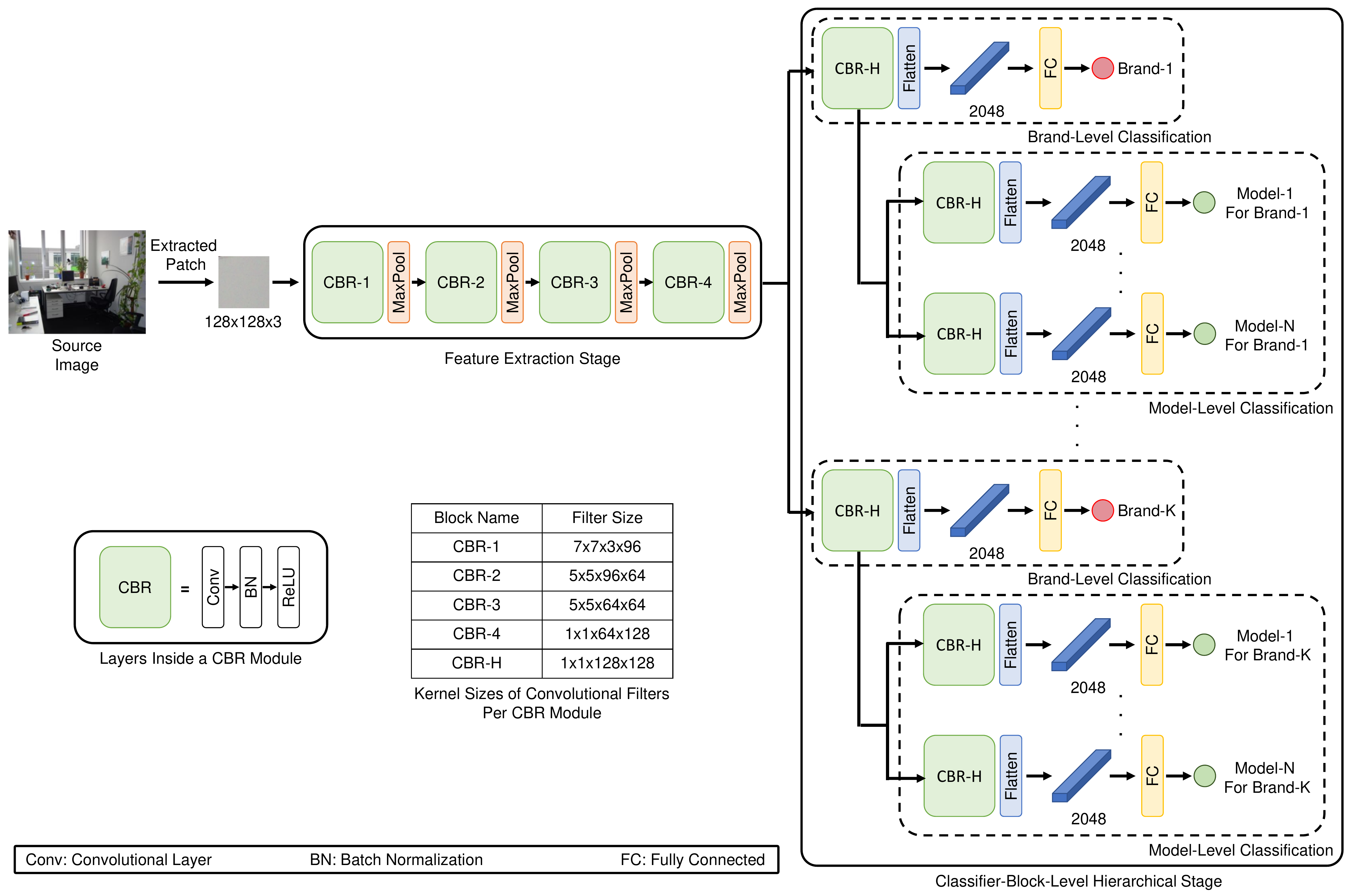}
 	\caption{The overall framework of the proposed method. A homogeneous patch of size 128x128 is first extracted from the source image. The patch is passed      to the feature extraction stage to extract the global feature map. This global feature map is fed to different brand-level and model-level branches           hierarchically for both camera brand and model identification.}
 	\label{proposed_framework}
\end{figure*}

Bondi \etal \cite{bondi} proposed a convolutional neural network-based approach for camera model identification to remove any analytical modeling. This allows the proposed model to be less prone to errors due to simplistic assumptions or model simplifications (e.g., linearizations, etc.). Tuama \etal \cite{tauma} utilized a denoising high-pass filter on the input image before passing it to the deep-learning framework. The filter helps by reducing the interference of image edges and textures better to identify the residuals responsible for camera model identification. However, this filter is predetermined and requires human intervention. To solve this, Bayer and Stamm \etal \cite{bayer_stamm} proposed a constrained convolutional layer that can adaptively learn the parameter of the filters. Marra \etal \cite{Marra2017} experimented with a set of local blind features based on co-occurrence matrices of selected neighbors and image residuals of different color bands. Rafi \etal \cite{Rafi2021} proposed a data-adaptive preprocessing block named remnant block that dynamically learns to suppress the scene details in an image for robust source camera identification. In their follow-up work \cite{Rafi_eccv}, they adopt an L2-constrained remnant block for learning a more robust preprocessing module. These methods, however, do not entirely prevent the ConvNets from learning the high-level scene details. Bennabhaktula \etal \cite{BENNABHAKTULA2022117769}  proposed a patch filtering approach by examining the homogeneity regions of the image that have more camera-specific traces rather than the scene details. Additionally, they adopted a hierarchical training pipeline that is more effective than a single classifier. This also facilitates the parallel training and addition of a new camera brand to the previously trained model. However, the network-level hierarchical approach substantially increases the amount of parameter and training cost. To address this, An alternative hierarchical approach is required to achieve the same performance level without substantially increasing the parameter and training cost.

\section{Methodology} 
The proposed ConvNet architecture is illustrated in Fig.~\ref{proposed_framework}. The initial feature extraction stage of the ConvNet network is adopted from the modified MISLNet proposed by Bennabhaktula \etal \cite{BENNABHAKTULA2022117769}. The feature extraction stage of the ConvNet architecture is divided into four blocks. We follow a similar pattern for all these blocks, where each block consists of a CBR module containing a convolution layer, batch normalization \cite{pmlr-v37-ioffe15}, and ReLU transformation. A max-pooling layer follows each CBR module layer. The kernel sizes of convolutional layers in each CBR module are given in Fig.~\ref{proposed_framework}.
\subsection{Classifier-Block-Level Hierarchical Module}
Let the feature map extracted from the initial stage is $\mathbf{X} \in \mathbb{R}^{C \times H \times W}$ and the number of brands present in our dataset is $K$. Then this global feature map is passed to $K$-number of different CBR modules. A separate CBR module for each brand allows the network to identify more discriminative features responsible for brand classification. This modular approach also facilitates newer brand addition. For example, if we want to add a new brand to the network, we can easily integrate another branch-level classification module, which will accept the global feature map $\mathbf{X}$. The feature map $\mathbf{X}$ is passed to the CBR module to generate feature space $\mathbf{X}_b \in \mathbb{R}^{C \times H \times W}$ for that particular brand. Next, we apply the flatten operation on $\mathbf{X}_b$ to generate the embedding $\mathbf{a}_b \in \mathbb{R}^{CHW}$ for that particular camera brand. This embedding is passed to a fully connected layer to generate a logit score for that particular camera brand.
\begin{equation}
    f_b^i = \mathbf{W}_b^i\mathbf{a}_b^i + b_b^i,\ i \in \{1,2,...,K\}
\end{equation}
Afterward, we apply the softmax operation to generate the probability score for that particular brand. We utilize the cross entropy loss for camera brand classification.
\begin{equation}
    p_b^i=\frac{\exp{(f_b^i)}}{\sum_{j=1}^{K}{\exp{(f_b^i)}}}\ ,  i \in \{1,...,K\}
\end{equation}
\begin{equation}
    \mathcal{L}_{bc}= -\frac{1}{K}\sum_{i=1}^{K}\bigg[l_b^i\log{\left(p_b^i\right)}+(1-l_b^i)\log{\left(1-p_b^i\right)}\bigg] 
\end{equation}
\begin{figure*}[!t]
	\centering
	\includegraphics[width=\linewidth]{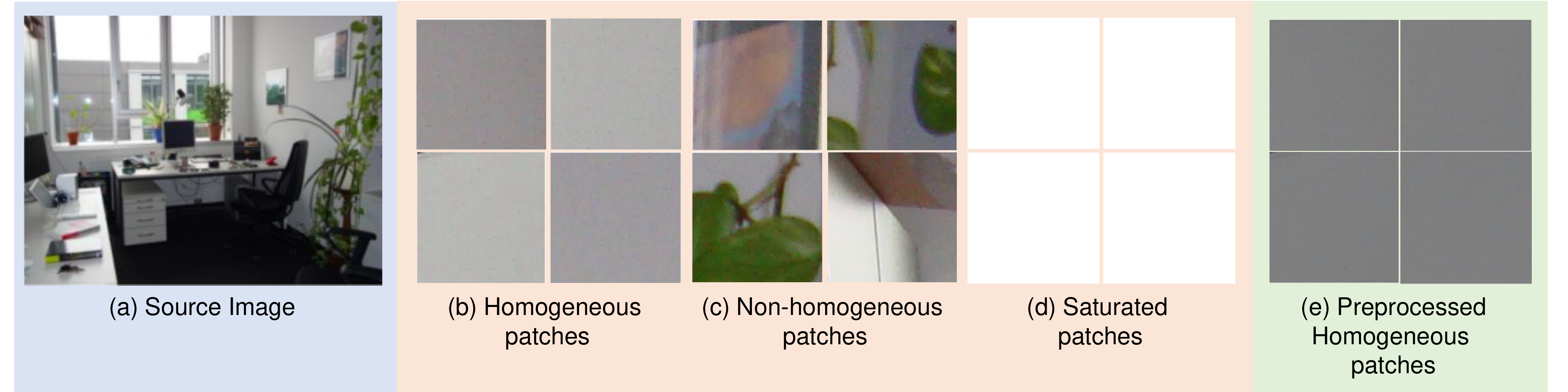}
 	\caption{An illustration of patch selection using standard deviation. (a) A sample image. (b) Examples of homogeneous patches (0.005 $\leq$ std $\leq$ 0.02), (c) nonhomogeneous patches (std $>$ 0.02), (d) saturated patches (std $<$ 0.005), (e) the corresponding pre-processed homogeneous patches after per-channel-mean subtraction.}
 	\label{patch_examples}
\end{figure*}
Here, $l_b^i$ represents the ground truth for a particular camera brand. Similarly for model-level classification, we break down the brand-level branch into hierarchical manner. Let the number of models per brand is $N$. Then we pass the feature space from brand-level $\mathbf{X}_b$ to $N$ number of CBR module to generate the camera model-level feature spaces $\mathbf{X}_m \in \mathbb{R}^{C \times H \times W}$ for that particular camera brand. Next, we flatten the feature space to generate the embedding $\mathbf{a}_m \in \mathbb{R}^{CHW}$ and pass through a FC layer to get the logit score. 
\begin{equation}
    f_m^i = \mathbf{W}_m^i\mathbf{a}_m^i + b_m^i,\ i \in \{1,2,...,N\}
\end{equation}
Finally, we apply softmax function and cross entropy loss for model classifcation.
\begin{equation}
    p_m^i=\frac{\exp{(f_m^i)}}{\sum_{j=1}^{N}{\exp{(f_m^i)}}}\ ,  i \in \{1,...,N\}
\end{equation}
\begin{equation}
    \mathcal{L}_{mc}= -\frac{1}{N}\sum_{i=1}^{N}\bigg[l_m^i\log{\left(p_m^i\right)}+(1-l_m^i)\log{\left(1-p_m^i\right)}\bigg] 
\end{equation}
Here, $l_m^i$ represents the ground truth for a camera model for a particular camera brand.
\subsection{Overall Objective Loss Function}
Our overall learning objective is comprised of the brand-level and model-level classification losses and is expressed as:
\begin{equation}
    \mathcal{L}_{total} = \mathcal{L}_{bc} + \alpha \mathcal{L}_{mc}
\end{equation}
The $\alpha$ is a hyperparameter and is determined through cross-validation experiments.

\subsection{Patch Selection}
We follow the method described in \cite{BENNABHAKTULA2022117769} for patch selection. Our goal is to extract patches from homogeneous regions with low-level scene details. These regions contain camera processing noise that is least distorted by high-level scene details. We tile the input image with blocks of 128x128 pixels, with a stride rate of $32\times32$ pixels. We avoid $256\times256$ pixels as they would result in fewer blocks. We also discard $64\times64$ blocks as extracting camera-specific features from smaller image patches is more complicated. Next, to calculate the homogeneity of each extracted block, we calculate the per-color channel standard deviation. If the standard deviation of all three channels is less than 0.02, we determine the block as homogeneous. To eliminate saturated patches, we set e lower threshold value of 0.005. Suppose we will select $p$ number of patches from an image. If the number of the homogeneous patches is less than the $p$, we choose the rest of the patches from the non-homogeneous and saturated patches. Finally, we subtract the per-channel mean from their respective channel to reduce the brightness effect. Fig. \ref{patch_examples} demonstrates an example image along with the homogeneous, non-homogeneous, and saturated patches extracted from the image.

\subsection{Data Balancing}
To handle the imbalance problem, we follow hierarchical data balancing from \cite{BENNABHAKTULA2022117769} instead of oversampling or under sampling. Let the total number of image patches that we would like to train per fold is $k$. We want to distribute the patches evenly across the training dataset. Let the number of camera brand is $n_{b}$. Then, we will select $k_{b}$ number of patches per camera brand by,
\begin{equation}
    k_{b} = [\frac{k}{n_{b}}]
\end{equation}
where $k, k_{b}, n_{b} \in N$, and $[.]$ denotes the standard rounding function. Let $n_{m}$ denotes the number of models representing the brand $b$. Let $k_{m}$ denotes the number of patches to be sampled from a particular model $m$. In order to keep the number of patches the same across different models of the same brand, $k_{m}$ can be set to:
\begin{equation}
    k_{m} = [\frac{k}{n_{m} \cdot n_{b}}]
\end{equation}
where $k_m, n_m \in N$. We continue this process to determine the number of patches to be sampled at the device-level $k_{d}$, for all the devices $n_d$ of model $m$. The value of $k_d$ is given by:
\begin{equation}
    k_{d} = [\frac{k}{n_d \cdot n_{m} \cdot n_{b}}]
\end{equation}
where $k_d, n_d \in N$. Finally, we determine the number of patches to be sampled from each image $i$ captured by the device $d$ as:
\begin{equation}
    k_{i} = [\frac{k}{n_i \cdot n_d \cdot n_{m} \cdot n_{b}}]
\end{equation}
where $n_i$ represents the number of images in the dataset belonging to the device $d$. Thus, for an initial estimate of $k$, the number of patches that need to be extracted from an image $i$ is determined by $k_i$, where $d$, $m$, and $b$ correspond to the device, model, and brand of the image,
respectively.

\section{Training}
In this section, we discuss the details of our training procedure. We also briefly discuss the dataset used for our experiments. All our experiments are performed in a hardware environment that includes an Intel Core-i7 7700k, @ 4.20 GHz CPU, and Nvidia GeForce GTX 1070 (8 GB Memory) GPU. All the necessary codes are written in Python, and we use Pytorch deep learning library \cite{NEURIPS2019_9015} to implement the neural networks.
\subsection{Dataset Description}
\begin{figure}[!t]
	\centering
	\includegraphics[width=\linewidth]{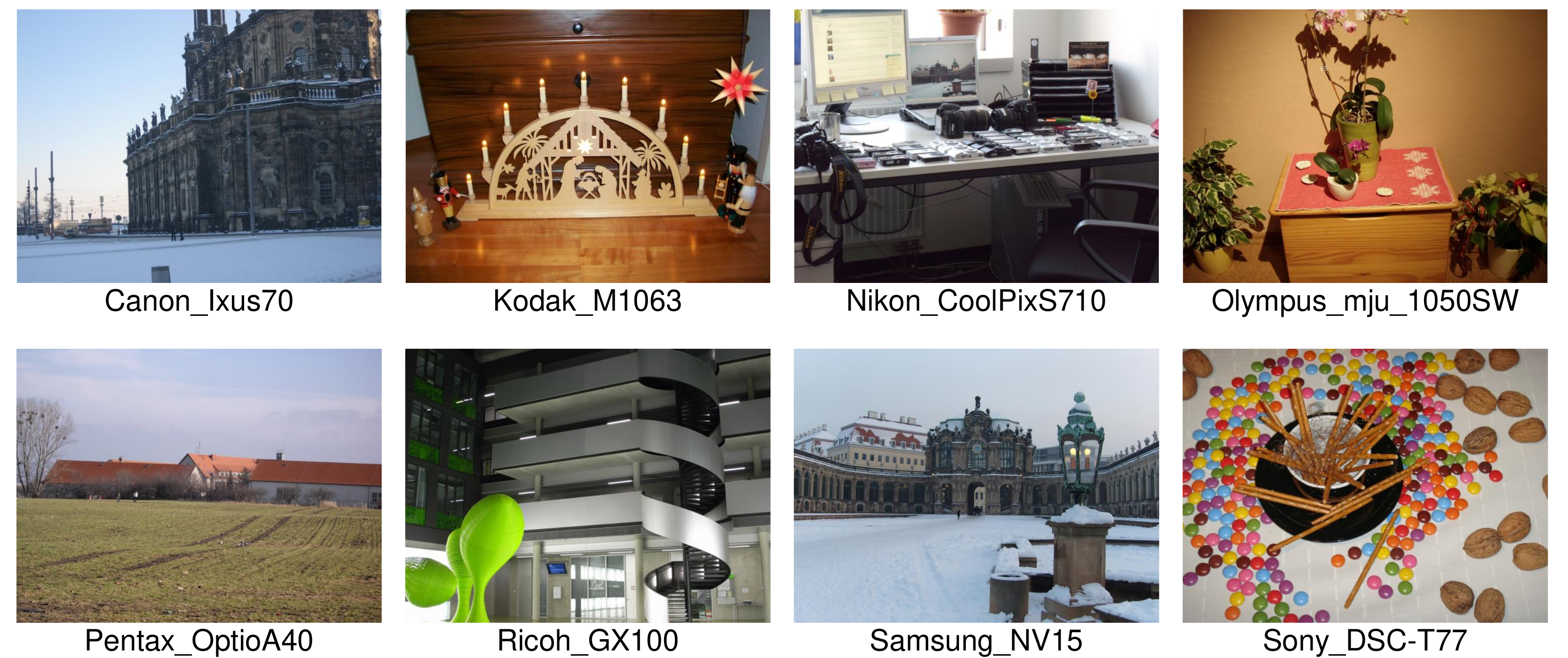}
 	\caption{An illustration of example images from 8 different camera model brands from the Dresden dataset.}
 	\label{dresden_img_examples}
\end{figure}

We utilize the publicly available Dresden dataset to evaluate our proposed framework \cite{dresden_dataset}. The images in the Dresden dataset are captured by 74 devices of 27 different camera models. The images are taken at several different locations using different views. The images are categorized into several subsets: namely JPEG, natural, flat-field, and dark-field, among others. The natural subset contains more diversified scenes than the other subsets. For that reason, following \cite{BENNABHAKTULA2022117769, Rafi2021}, we use the natural subset for our experimentation. We consider only those camera models that have at least two devices for that camera model. This enables us to keep one device per camera model aside for testing while using the rest for training. Out of the 27 camera models, 19 models meet this criterion. As suggested by \cite{nikon_merge}, we merge the camera model Nikon\_D70 and Nikon\_D70s into a single class named Nikon\_D70. Finally, the dataset represents 18 camera models and 66 individual camera instances. A brief description regarding the number of images and number of devices per camera model is given in Table \ref{Table: Dresden Dataset Details}. A few example images from different camera model brands from the Dresden dataset are shown in Fig. \ref{dresden_img_examples}.

\begin{table}[!t]
    \centering
    \caption{\textsc{Details of the Camera Models from the Dresden Dataset Used in Our Experiments.}}
    \label{Table: Dresden Dataset Details}
    \begin{tabular}{cccc}
    \toprule
    Serial No & Camera model name & No. of devices & No. of images \\ 
    \midrule
    1 & Canon\_Ixus70 & 3 & 603 \\
    2 & Casio\_EX-Z150 & 5 & 964 \\
    3 & FujiFilm\_FinePixJ50 & 3 & 661 \\
    4 & Kodak\_M1063 & 5 & 2527 \\
    5 & Nikon\_CoolPixS710 & 5 & 961 \\
    6 & Nikon\_D200 & 2 & 1697 \\
    7 & Nikon\_D70 & 4 & 1664 \\
    8 & Olympus\_mju\_1050SW & 5 & 1064 \\
    9 & Panasonic\_DMC-FZ50 & 3 & 988 \\
    10 & Pentax\_OptioA40 & 4 & 760 \\
    11 & Praktica\_DCZ5.9 & 5 & 1056 \\
    12 & Ricoh\_GX100 & 5 & 1059 \\
    13 & Rollei\_RCP-7325XS & 3 & 625 \\
    14 & Samsung\_L74wide & 3 & 720 \\
    15 & Samsung\_NV15 & 3 & 676 \\ 
    16 & Sony\_DSC-H50 & 2 & 630 \\
    17 & Sony\_DSC-T77 & 4 & 777 \\
    18 & Sony\_DSC-W170 & 2 & 439 \\
    \bottomrule
    \end{tabular}
\end{table}

\subsection{Implementation Details}
We set the parameter $k$ to 260000 for data balancing. We utilize the stochastic gradient descent optimizer with a learning rate of 0.1 and momentum of 0.9. We schedule the learning rate with an exponential learning rate decay with a factor of 0.9. We use L2 regularization with a weight decay factor of 0.005. We set the batch size to 512. We run the training for 40 epochs and select the weight with the highest classification accuracy on the validation set for inferencing on the test dataset. During inference, we select 200 patches per image and make image-level predictions. For making image-level predictions, we perform majority voting. If the model predicts a brand with multiple camera models, we use the respective model-level branch to identify the camera model. There are a total of 13 camera brands and 2 unique camera models in the case of the Samsung brand, 3 in the case of the Nikon and Sony brand, in the Dresden dataset. For evaluation purposes, we have selected classification accuracy as the evaluation metric.

\begin{table*}[!t]
    \centering
    \caption{\textsc{Comparison with the State-of-the-art on the ``Natural" Subset of the Dresden Image Dataset for Source Camera Model Identification.}}
    \begin{adjustbox}{width=\textwidth}
    \label{Table: result comparison}
    \begin{tabular}{cccc}
    \toprule
    Method & Accuracy & No. of Models & Evaluation schemes \\ 
    \midrule
    Tuama \etal \cite{tauma} & 0.9709 & 14 & 5-fold CV \\
    Marra \etal \cite{Marra2017} & 0.9627 & 25 & 20-fold CV with 7 devices used for both train and test sets \\
    Marra \etal \cite{Marra2017} & 0.9872 & 25 & 20-fold CV with 7 devices used for both train and test sets \\
    Bondi \etal \cite{bondi} & 0.9650 & 18 & Scene independent test set \\
    Rafi \etal \cite{Rafi2021} & 0.9703 & 18 & Scene independent test set \\
    Rafi \etal \cite{Rafi_eccv} & 0.9815 & 18 & Scene independent test set\\
    Bennabhaktula \etal \cite{BENNABHAKTULA2022117769} - 200 patches & 0.9899 & 18 & Leave-one-device-out 5-fold CV \\
    Ours - 200 patches & 0.9903 & 18 & Leave-one-device-out 5-fold CV \\
    \bottomrule
    \end{tabular}
    \end{adjustbox}
\end{table*}

\begin{table*}[!t]
    \centering
    \caption{\textsc{Network Parameter-number Comparison with the State-of-the-art Hierarchical Method.}}
    \label{Table: parameter comparison}
    \begin{tabular}{cccc}
    \toprule
    Method & Network's Total Parameter Number \\ 
    \midrule
    Bennabhaktula \etal \cite{BENNABHAKTULA2022117769} & 2,585,149 \\
    Ours & 674,517 \\
    \bottomrule
    \end{tabular}
\end{table*}

\begin{table*}[!t]
    \centering
    \caption{\textsc{Comparison of the Classification Accuracy Between the Hierarchical and the Flat Approaches.}}
    \label{Table: Hierarchical vs flat approach result comparison}
    \begin{tabular}{ccccccc}
    \toprule
    Classification & fold-1 & fold-2 & fold-3 & fold-4 & fold-5 & average \\ 
    \midrule
    Flat & 0.9801 & 0.9903 & 0.9843 & 0.9824 & 0.9881 & 0.9850 $\pm$ 0.0037 \\
    Hierarchical & 0.9904 & 0.9914 & 0.9893 & 0.9908 & 0.9897 & 0.9903 $\pm$ 0.0008 \\     
    \bottomrule
    \end{tabular}
\end{table*}

\section{Experimental Results And Analysis}
\subsection{Comparison With State-of-the-Arts}
\begin{figure}[!t]
	\centering
	\includegraphics[scale=0.5]{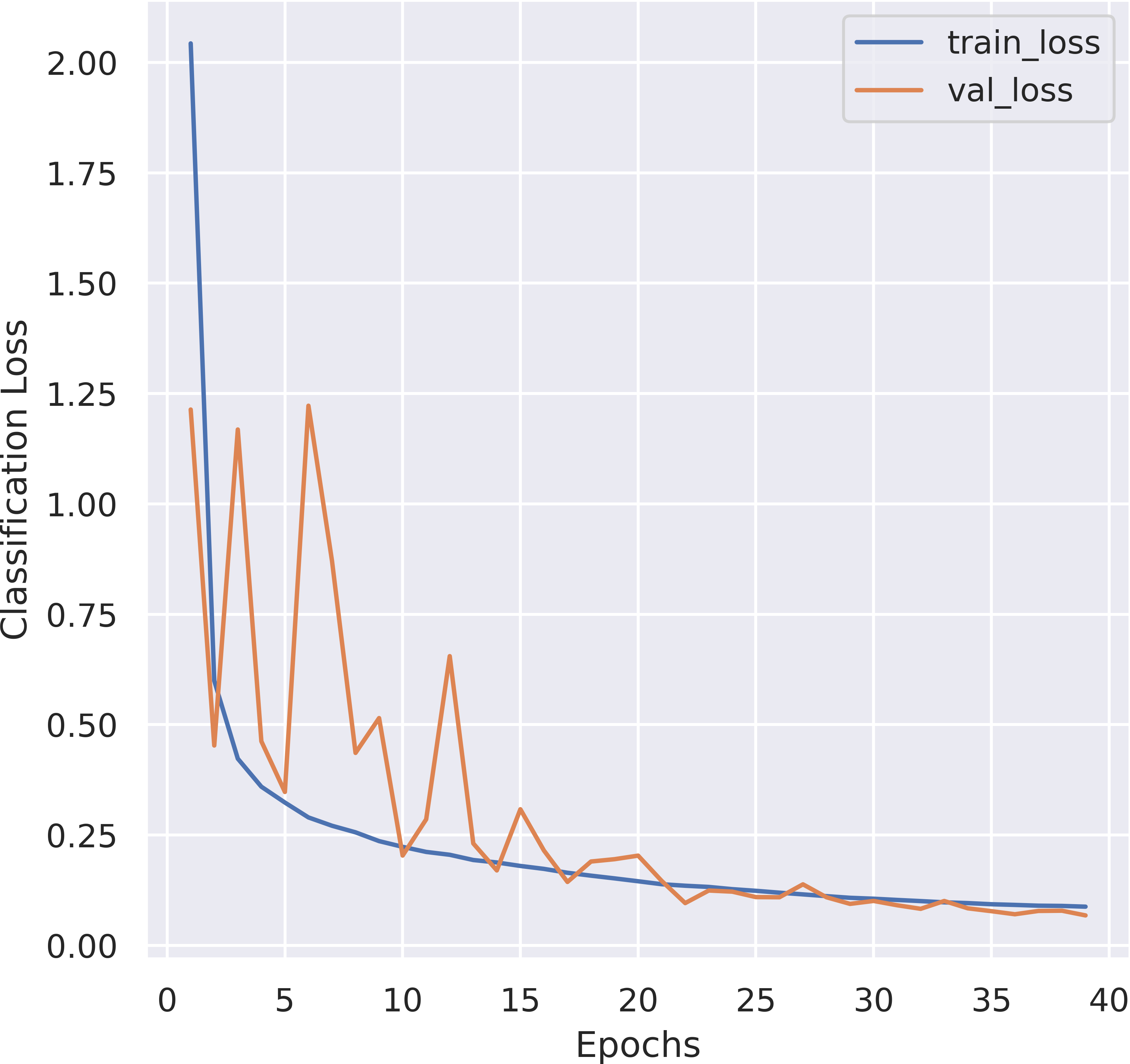}
 	\caption{The plots depict the convergence of the loss for fold 1 (out of 5).}
 	\label{loss_curve}
\end{figure}

To evaluate the proposed method, we perform five-fold cross-validation. We leave out one device from each camera brand and model for testing each fold. The camera brands have an unequal number of devices. Due to that, we rotationally select one device and leave it out for testing purposes. Fig. \ref{loss_curve} shows the convergence in the loss for fold-1, achieved with the hyper-parameters mentioned above. We compare our proposed framework with previously published state-of-the-art methods including: the methods of Tuama \etal \cite{tauma}, Marra \etal \cite{Marra2017}, Bondi \etal \cite{bondi}, Rafi \etal \cite{Rafi2021}, Rafi \etal \cite{Rafi_eccv}, and Bennabhaktula \etal \cite{BENNABHAKTULA2022117769}. The results have been reported from their implementations and are shown in Table \ref{Table: result comparison}. We also compute the total parameter number of the state-of-the-art network-level hierarchical method of Bennabhaktula \etal \cite{BENNABHAKTULA2022117769} and compare it with ours. We use the torch-summary \footnote{\url{https://pypi.org/project/torch-summary/}} library to calculate the parameter number. The comparison between parameter numbers is given in Table \ref{Table: parameter comparison}. As shown in Table \ref{Table: result comparison} and \ref{Table: parameter comparison}, our proposed classifier-block-level-based hierarchical network can achieve a similar level of classification performance (accuracy of 0.9903) compared to the previous state-of-the-art network-level hierarchical method (0.9899). However, our proposed model requires only 674,517 parameters compared to the previous state-of-the-art method's 2,585,149 parameters. Our proposed framework can achieve a similar level of performance but requires 74\% fewer parameters.

\subsection{Hierarchical vs flat approach}
In order to evaluate the effect of the hierarchical approach compared to the flat approach, we have conducted an experiment where we trained all 18 camera models using a single classifier. We flatten the global feature map $\mathbf{X}$, and feed it to three fully connected layers sequentially for classification. The results are given in Table \ref{Table: Hierarchical vs flat approach result comparison}. From table \ref{Table: Hierarchical vs flat approach result comparison}, it can be seen that the hierarchical framework performs better compared to the flat approach.

\subsection{Future work}
One direction for future research is to extend this work for device-level identification per camera model. This will be particularly helpful when it is necessary to discriminate between two separate devices of the same camera model and brand. Another approach would be to design a loss function that can explicitly exploit the hierarchical relationship between a brand-level branch and the model-level branches for that particular brand. It can further improve the classification performance as model-level semantic features are embedded in the brand-level semantic features.

\section{Conclusion}
We propose a classifier-block-level-based hierarchical classification approach for source camera model identification. This approach significantly reduces the required number of parameters but can retain the same level of performance compared to a state-of-the-art network-level hierarchical approach. Our proposed approach is also modular in nature. As a result, newer camera brands or models can be easily integrated into the framework. Our experiments also show that our proposed method achieves state-of-the-art performance on the Dresden dataset.

\balance
\bibliographystyle{IEEEtran}
\bibliography{IEEEabrv,main.bib}

\end{document}